\documentclass{bmvc2k}

\title{\textbf{PEEK}: \textbf{P}icking \textbf{E}ssential frames via \textbf{E}fficient \textbf{K}nowledge distillation}

\addauthor{Killian Steunou}{killian.steunou@ip-paris.fr}{1,2}
\addauthor{Anas Filali Razzouki}{anas@momentslab.com}{1,2}
\addauthor{Khalil Guetari}{khalil.guetari@momentslab.com}{2}
\addauthor{Mounîm A. El-Yacoubi}{mounim.el_yacoubi@telecom-sudparis.eu}{1}
\addauthor{Yannis Tevissen}{yannis.tevissen@momentslab.com}{2}

\addinstitution{
 Télécom SudParis — SAMOVAR\\
 Institut Polytechnique de Paris\\
 Palaiseau, France
}
\addinstitution{
 Moments Lab\\
 69 Avenue Pierre Grenier\\
 Boulogne-Billancourt, France
}

\runninghead{Steunou \etal}{PEEK}

\usepackage{adjustbox}

\def\etal{\emph{et al}\bmvaOneDot}

\begin{document}

\maketitle

\begin{abstract}
Video-language models can process only a limited number of frames, making frame selection a key bottleneck for efficient video captioning. Most captioning pipelines still rely on uniform sampling, which is computationally cheap but agnostic to visual content. Adaptive frame sampling selects the most informative frames from a video, but existing methods remain computationally expensive. We introduce PEEK, an efficient dynamic frame sampling method that distills caption-conditioned frame relevance rankings from a stronger teacher model into a lightweight temporal model that operates only on visual content.
On ActivityNet Captions and MSR-VTT, PEEK outperforms state-of-the-art methods across all evaluated downstream vision language models, obtaining the best CIDEr for most frame budgets, especially when only one or two frames are selected, winning 14 out of 16 configurations on ActivityNet Captions. Zero-shot on MSR-VTT, it transfers best at low frame budgets, while results at four and eight frames are more mixed as temporal coverage and visual diversity become increasingly competitive. Compared with recent adaptive baselines, PEEK is both more accurate in the low-budget regime and more efficient: it adds only $5.2\%$ to the captioning time, compared with $65.4\%$ for CSTA and $211.9\%$ for MaxInfo. We release our code and pre-trained checkpoint at \href{https://github.com/momentslab/peek}{https://github.com/momentslab/peek}.
\end{abstract}

\section{Introduction}
\label{sec:intro}

\begin{figure}[!htbp]
    \centering
    \begin{minipage}[t]{0.5\textwidth}
        \vspace{0pt}
        \begin{subfigure}{\linewidth}
            \centering
            \includegraphics[width=\linewidth]{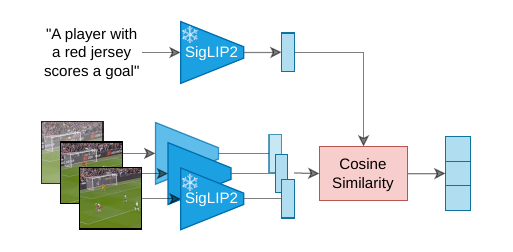}
            \caption{Stage~1 --- Oracle teacher scoring}
            \label{fig:pipeline-phase1}
        \end{subfigure}
    \end{minipage}\hfill
    \begin{minipage}[t]{0.5\textwidth}
        \vspace{0pt}
        \begin{subfigure}{\linewidth}
            \centering
            \includegraphics[width=\linewidth]{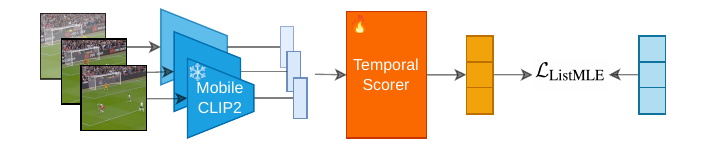}
            \caption{Stage~2 --- query-free temporal scorer}
            \label{fig:pipeline-phase2}
        \end{subfigure}

        \vspace{8pt}

        \begin{subfigure}{\linewidth}
            \centering
            \includegraphics[width=1\linewidth]{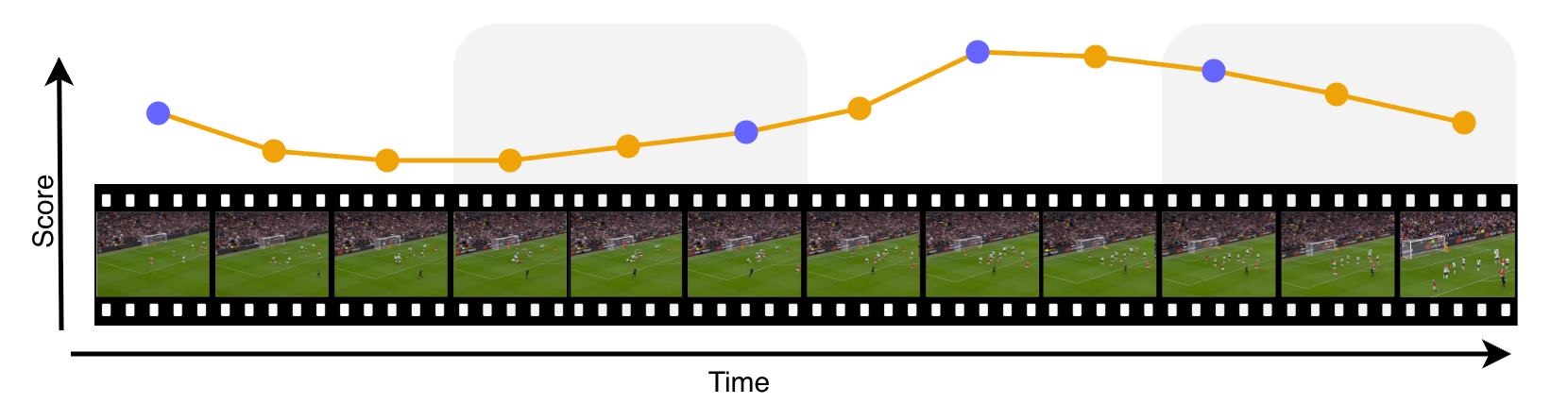}
            \caption{Stratified argmax inference with $k=4$}
            \label{fig:pipeline-inference}
        \end{subfigure}
    \end{minipage}
    \vspace{1pt}
    \caption{%
    \textbf{Overview of PEEK.} \textbf{(a)}~A frozen SigLIP\,2 dual encoder acts as an Oracle teacher, producing per-frame relevance targets from ground-truth captions. \textbf{(b)}~A small Transformer distills the teacher's ranking into a query-free selector operating on MobileCLIP2 visual embeddings alone. \textbf{(c)}~At inference, the segment is split into $k$ equal temporal windows and the highest-scoring frame within each (blue dot) is kept.}
    \label{fig:pipeline}
\end{figure}

Modern vision--language models (VLMs) have made strong progress on image-language tasks, but video understanding remains expensive because videos are long, redundant, and often require sparse relevant cues to be extracted from a frame sequence~\cite{tangVideoUnderstandingLarge2023, wuLongVideoBenchBenchmarkLongcontext2024}. To process a video, a VLM usually receives only a limited number of frames: enough to give a glimpse of the video, but not enough to guarantee that the decisive visual cue is present. In practice, even for state-of-the-art models, the default strategy is uniform sampling: partition the video into equal temporal segments and keep one frame from each~\cite{zoharApolloExplorationVideo2024, baiQwen25VLTechnicalReport2025, marafiotiSmolVLMRedefiningSmall2025, baiQwen3VLTechnicalReport2025a}. Uniform sampling is deterministic, model-free, and often produces good results~\cite{huMLLMBasedVideo2025, yuFrameVoyagerLearningQuery2025, huangAdaptiveGreedyFrame2026}, outperforming adaptive strategies on some benchmarks~\cite{brkicFrameSamplingStrategies2025}. This makes uniform sampling a good baseline rather than a naive approach. However, it is still fundamentally content-blind: a short clip where the key event happens in a single instant and a clip where useful evidence is spread across the whole duration are treated identically~\cite{huMLLMBasedVideo2025,huangAdaptiveGreedyFrame2026}.
Other works use a strong text-conditioned retriever, such as CLIP \cite{radfordLearningTransferableVisual2021a} or SigLIP \cite{zhaiSigmoidLossLanguage2023}, to score every frame by image--text similarity and keep the most relevant frames \cite{sunFramesClipsTrainingfree2025}. This process is query-dependent, so it cannot be used at inference time for video captioning, but it can diagnose how much visually relevant frames improve captioning. We therefore propose a caption-conditioned Oracle as a diagnostic for frame relevance: it scores candidate frames against the ground-truth caption and ranks them by semantic alignment with the target description. Its rankings provide a useful supervisory signal, indicating which frames are visually salient or semantically aligned with the caption, and may indirectly reveal temporally distinctive moments within the segment. Our hypothesis is that part of this caption-conditioned relevance can be distilled into a lightweight visual model that never sees text at inference time.\\

Concretely, we propose PEEK, a two-stage distillation framework for efficient frame selection in video captioning: a frozen vision-language teacher first scores candidate frames against the ground-truth caption, producing dense relevance rankings used only for supervision, and a lightweight temporal Transformer then learns to predict these rankings from visual embeddings alone. At inference time PEEK requires neither captions nor a text encoder. The deployed selector is query-free, caption-agnostic, and independent of the downstream captioning model.

We make the following contributions:
\begin{itemize}
    \item We introduce PEEK, a query-free frame selector that distills SigLIP\,2~\cite{tschannenSigLIP2Multilingual2025} caption-conditioned rankings into a lightweight temporal scorer operating only on visual features.
    \item We propose caption-conditioned frame scoring as an Oracle diagnostic to quantify the value of semantic frame relevance for video captioning.
    \item We evaluate PEEK on ActivityNet Captions~\cite{krishnaDenseCaptioningEventsVideos2017a} and MSR-VTT~\cite{xuMSRVTTLargeVideo2016} with four downstream VLMs, showing consistent gains in the low-frame regime and a much lower selection cost than recent content-aware baselines.
\end{itemize}

The paper is organized as follows. Section~\ref{sec:related} reviews prior work on frame selection. Section~\ref{sec:method} presents PEEK, Section~\ref{sec:experiments} describes the datasets, experimental protocol, and results. Finally, we discuss the limitations of our approach and conclude the paper.

\section{Related Work}
\label{sec:related}

VLMs usually operate under a fixed visual-token or frame budget, which makes temporal sampling a central design choice rather than a neutral preprocessing step. Uniform sampling remains widely used because it is deterministic, model-free, and cheap --- and it is a strong baseline: Brkic \etal~\cite{brkicFrameSamplingStrategies2025} show, in a controlled benchmark for small VLMs, that frame-sampling choices substantially affect video question-answering results, with uniform sampling the strongest strategy on VideoMME~\cite{fuVideoMMEFirstEverComprehensive2025}. Adaptive selectors must therefore be compared against it carefully. Uniform sampling nevertheless ignores the large variation in information density across videos. 

\paragraph{Training-free frame selection.} Training-free methods replace uniform sampling with adaptive keyframe selection, typically optimizing informativeness, diversity, and temporal coverage. MaxInfo selects representative frames by maximizing the geometric volume spanned by frame embeddings~\cite{liMaxInfoTrainingFreeKeyFrame2025}. Other adaptive samplers combine text relevance with visual coverage~\cite{tangAdaptiveKeyframeSampling2025,huangAdaptiveGreedyFrame2026,zhangQFrameQueryawareFrame2025,tanThinkClipSampleSlowFastFrame2026}, which makes them query-dependent. All require a large visual encoder pass over densely sampled frames, making them computationally expensive.

\paragraph{Learned frame selectors.}
Frame-Voyager~\cite{yuFrameVoyagerLearningQuery2025} ranks candidate frame combinations by the prediction loss of a pretrained Video-LLM. M-LLM-based frame selection~\cite{huMLLMBasedVideo2025} trains a lightweight multimodal selector from M-LLM and LLM pseudo-labels covering single-frame importance and multi-frame temporal signals. VideoBrain~\cite{zouVideoBrainLearningAdaptive2026} treats sampling as an adaptive acquisition process in which a VLM invokes retrieval or local dense-sampling agents on demand. Video summarization methods such as CSTA~\cite{sonCSTACNNbasedSpatiotemporal2024} also learn frame-importance scores, although for a different objective. Frame selection is thus increasingly treated as a learnable decision process, but these selectors are often tied to the downstream model or input query.

\paragraph{Frame selection for video captioning.}
Video captioning is a particularly constrained case for adaptive sampling: only the visual signal is available at selection time, so text-aware selectors cannot be applied directly. PickNet~\cite{chenLessMorePicking2018} selects compact frame subsets with reinforcement learning and task-specific rewards; LFS~\cite{chaoLFSLearnableFrame2026} learns a selector for detailed captioning from caption feedback produced by frozen video-LLMs, and despite few learnable parameters still relies on an expensive vision-language backbone. Both share our motivation --- fewer frames at preserved caption quality --- but differ in how frame importance is supervised.

\paragraph{Text-conditioned frame scoring.}
Dual-encoder models such as CLIP~\cite{radfordLearningTransferableVisual2021a} and SigLIP~\cite{zhaiSigmoidLossLanguage2023} make it natural to score frames against a caption or question through image--text similarity, used for video retrieval, keyframe selection, and data management, such as KeyVideoLLM~\cite{liangKeyVideoLLMLargescaleVideo2024} for large-scale keyframe selection and VideoLLM data compression. Such scoring estimates well which frames align with a sentence, but requires the text before selection and a dense encoder pass over many candidates. Ranking-based objectives are a natural fit, since only the relative order of frames determines which visual evidence reaches the downstream model~\cite{xiaListwiseApproachLearning2008,plackettAnalysisPermutations1975}.

\paragraph{When does frame selection matter?}
TempCore~\cite{okTempCoreAreVideo2026} introduces Frame Selection Sensitivity and reports that many video question-answering examples are largely frame-agnostic, with only a subset genuinely sensitive to which frames are shown. Gains from selection should therefore not be expected uniformly across datasets, budgets, and downstream models: learned selection is most likely to help when the visual budget is tight or the relevant evidence is sparse, while uniform coverage strengthens as more frames are allowed.

\paragraph{Positioning of PEEK.}
PEEK shares the motivation of PickNet and LFS but differs in its supervision and deployment cost: rather than learning frame importance from captioner feedback or running a large vision-language model at selection time, it distills Oracle caption--frame relevance rankings into a small visual temporal scorer, using the text-conditioned model only to generate supervision. Unlike diversity-based methods such as MaxInfo, PEEK does not optimize visual diversity directly but learns a caption-oriented relevance prior.

\section{Method}
\label{sec:method}

We propose a two-stage framework for learning query-free temporal frame selectors. In Stage~1, a strong text-conditioned vision-language teacher scores every candidate frame of a video segment against its ground-truth caption, producing per-frame relevance signals. In Stage~2, a lightweight temporal scorer is trained to imitate the induced ranking without access to the caption. At inference time it can score each frame from an unseen video to select high-scoring frames before caption generation. Figure~\ref{fig:pipeline} summarizes our method.

\subsection{Stage~1: Oracle Scoring}
\label{sec:stage1}

Given an annotated temporal segment $(v, [t_s, t_e], c)$, where $v$ is the source video, $[t_s, t_e]$ is the temporal window, and $c$ is the associated caption, we subsample candidate frames from this window, resulting in a set of $T$ frames denoted $\mathcal{F} = \{f_1, \dots, f_T\}$. The reason is that video segments are long and redundant and it is typical in the state-of-the-art to subsample the video before applying a frame selection algorithm.
For each training segment we compute a per-frame relevance score using a frozen text-conditioned vision-language teacher. We use SigLIP\,2 as the teacher: frames are processed by its vision encoder, while the caption is processed by its text encoder. Finally, each frame is assigned a relevance score based on the cosine similarity between its visual embedding and the caption textual embedding. Embeddings are L2-normalized prior to cosine similarity computation.

Let $\psi_v$ denote the teacher vision encoder and $\psi_t$ the teacher text encoder.
Given teacher frame embeddings $\mathbf{z}_t = \psi_v(f_t) \in \mathbb{R}^{d_S}$ and the caption embedding $\mathbf{u} = \psi_t(c) \in \mathbb{R}^{d_S}$ with $d_S$ being the embedding dimension of SigLIP\,2, the raw teacher score for frame $t$ is the cosine similarity
\begin{equation}\label{eq:cosine}
    s_t = \frac{\langle \mathbf{z}_t,\, \mathbf{u} \rangle}{\lVert \mathbf{z}_t \rVert \, \lVert \mathbf{u} \rVert}.
\end{equation}
Only the resulting scalar scores are retained as supervision for the student. The caption embedding and teacher visual embeddings are not used as Stage~2 inputs.
The vector $\mathbf{s} = (s_1, \dots, s_T)$ is transformed into a training target. We min--max rescale each $s_t$ to obtain the final score $y = (y_1, \dots, y_T)$ which bounds targets to $[0, 1]$ while preserving the teacher's internal ordering.

\subsection{Stage~2: Caption-Agnostic Temporal Scorer}
\label{sec:scorer}

The student model is trained on embeddings obtained from a frozen lightweight MobileCLIP2~\cite{faghriMobileCLIP2ImprovingMultiModal2025}, while SigLIP\,2 is used only to produce the supervision targets. Let \(\varphi_v\) be the frozen vision encoder and let
\begin{equation}
    \mathbf{x}_t = \varphi_v(f_t) \in \mathbb{R}^{512},
    \qquad t=1,\dots,T,
\end{equation}
be the visual embedding of frame \(f_t\). Given the sequence
$\mathbf{X}=(\mathbf{x}_1,\dots,\mathbf{x}_T)$, we compute
\begin{equation}
    \mathbf{H}^{(0)} 
    =
    [\mathbf{h}_1^{(0)},\dots,\mathbf{h}_T^{(0)}]^\top,
    \qquad
    \mathbf{h}_t^{(0)}
    =
    W_{\mathrm{in}}\mathrm{LN}(\mathbf{x}_t) + \mathbf{p}_t + \mathbf{b}_{\mathrm{in}},,
\end{equation}
\begin{equation}
    \mathbf{H}^{(\ell)}
    =
    \mathrm{TransformerLayer}^{(\ell)}
    \left(\mathbf{H}^{(\ell-1)}\right),
    \qquad \ell=1,\dots,L,
\end{equation}
\begin{equation}
    \hat y = [\hat y_1, \dots, \hat y_T], \qquad
    \hat y_t = \mathbf{w}^{\top}\mathbf{h}_t^{(L)} + b.
\end{equation}
Here, $h$ is the hidden dimension, $W_{\mathrm{in}}\in\mathbb{R}^{h\times512}$, $\mathbf{b}_{\mathrm{in}}\in\mathbb{R}^{h}$, $\mathrm{LN}$ denotes layer normalization, $\mathbf{p}_t\in\mathbb{R}^{h}$ is a fixed sinusoidal positional encoding, $\mathbf{H}^{(\ell)}\in\mathbb{R}^{T\times h}$, $\mathbf{w}\in\mathbb{R}^{h}$, and $b\in\mathbb{R}$. The scalar \(\hat y_t\) is an unconstrained relevance logit for frame \(t\). The model has $L$ encoder layers with multi-head self-attention, ReLU-activated feed-forward blocks, and dropout. Our model has about 1.7M trainable parameters, excluding the frozen MobileCLIP2-S0 encoder. In total, it has only 13.1M parameters.

Pointwise regression on $(y_t, \hat y_t)$ pairs ignores the fact that frame selection is fundamentally a ranking problem: only the order among candidate frames affects downstream selection. We therefore use the ListMLE listwise objective of Xia \etal~\cite{xiaListwiseApproachLearning2008}. Let \(\pi^*\) be the permutation of frame indices sorted by decreasing teacher target, such that \(\pi^*(r)\) is the index of the frame at rank \(r\), and
\begin{equation}
y_{\pi^*(1)} \geq y_{\pi^*(2)} \geq \dots \geq y_{\pi^*(T)}.
\end{equation}
Viewing the model outputs $\hat{\mathbf{y}}$ as Plackett--Luce utilities \cite{plackettAnalysisPermutations1975}, the negative log-likelihood of observing $\pi^*$ under the model is
\begin{equation}\label{eq:listmle}
    \mathcal{L}_{\text{ListMLE}} = -\sum_{t=1}^{T} \log \frac{\exp\bigl(\hat y_{\pi^*(t)}\bigr)}{\sum_{\tau=t}^{T} \exp\bigl(\hat y_{\pi^*(\tau)}\bigr)}.
\end{equation}
Unlike pointwise MSE or pairwise hinge losses, this objective optimizes the probability of the teacher-induced ranking, aligning the training signal with the selection problem.

\subsection{Inference-Time Frame Selection}
\label{sec:selection}

At test time we score all candidate frames of an unseen segment at once, and select a budget of $k$ frames. Motivated by empirical measurements, we use a simple stratified argmax rule to select frames. We partition the segment into $k$ non-overlapping temporal sub-segments and select the highest-scoring frame inside each sub-segment:
\begin{equation}\label{eq:stratified}
    \mathcal{B}_j =
    \left\{
    \left\lfloor \frac{(j-1)T}{k} \right\rfloor + 1,\,
    \dots,\,
    \left\lfloor \frac{jT}{k} \right\rfloor
    \right\},
    \qquad j=1,\dots,k,
\end{equation}
\begin{equation}
    t_j^* = \arg\max_{t \in \mathcal{B}_j} \hat y_t,
    \qquad
    \mathcal{S}_k = (t_1^*, \dots, t_k^*).
\end{equation}
This policy combines two priors: the scorer chooses content-rich frames locally, while the sub-segments preserve temporal coverage. For $k=1$, stratified argmax reduces to selecting the single highest-scoring frame in the video. The selected frames are sorted in temporal order before being forwarded to the downstream captioning model.

\subsection{Implementation Details}
\label{sec:impl}

PEEK is trained on ActivityNet Captions (ANC) train segments only. SigLIP\,2 \\(\texttt{so400m-patch14-384}) is used to precompute teacher targets, while MobileCLIP2-S0 is used to precompute the frozen 512-dimensional student inputs. The temporal scorer uses hidden size $h = 256$, $L = 2$ encoder layers, $4$ attention heads, feed-forward dimension $1024$, and dropout $0.15$. We train with ListMLE, AdamW~\cite{loshchilov2018decoupled} with learning rate $2 \times 10^{-4}$ and a cosine annealing schedule, weight decay $0.03$, batch size $1024$, $25$ epochs, $2$ warmup epochs, and gradient clipping at $\lVert g \rVert_2 = 1.0$. Training uses light temporal augmentation: random frame drop in $[0.05, 0.25]$ and random crop with minimum fraction $0.7$. For efficient training, sequences are capped at $32$ frames per segment, with at least $6$ frames retained after augmentation.

\section{Experiments}
\label{sec:experiments}

\begin{figure*}[!htbp]
  \centering
  \includegraphics[width=\linewidth]{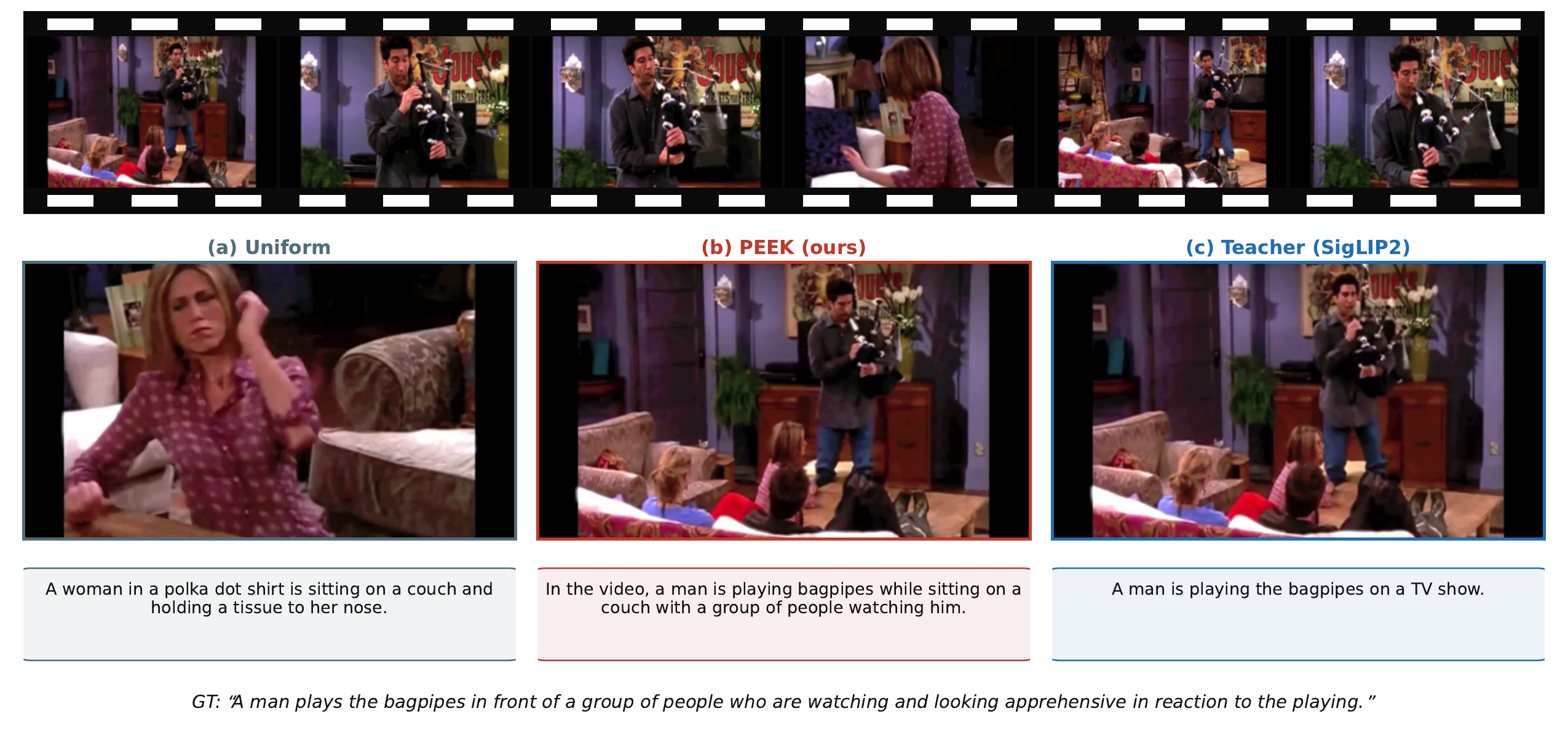}
  \caption{%
    Top frames selected on an ActivityNet Captions test segment in which a man plays the bagpipes in front of an audience. From left to right: (a)~the uniform (center) frame, (b)~the top-ranked frame from PEEK, and (c)~the top-ranked frame from the SigLIP2 teacher. The one sentence caption below each frame is generated by Qwen2.5-VL-3B; the ground-truth caption is shown at the bottom. Both PEEK and the teacher find the instrument, while the central frame misses it.%
  }
  \label{fig:qual_selection}
\end{figure*}

\subsection{Data}
\label{sec:data}

We train our model on ActivityNet Captions~\cite{krishnaDenseCaptioningEventsVideos2017a}, using the official splits and report all metrics on the test set. We also evaluate on the MSR-VTT~\cite{xuMSRVTTLargeVideo2016} test split to assess zero-shot transfer to clip-level captioning. Full split statistics are provided in the supplementary material.

\paragraph{ActivityNet Captions.}
ANC consists of untrimmed YouTube videos drawn from the
ActivityNet dataset~\cite{cabaheilbronActivityNetLargeScaleVideo2015}, each densely annotated with multiple temporally-localized natural-language descriptions. A single video contains, on average, between 3 and 4 overlapping or sequential events, with a typical total duration of about two minutes; every annotated event is described by a free-form English sentence. We train on 37{,}421 segments from 10{,}009 videos and evaluate on the 17{,}505 test segments from 4{,}885 videos (average segment duration ${\sim}40$\,s).

\paragraph{MSR-VTT.}
MSR-VTT contains short web video clips paired with 20 crowd-sourced English captions per clip. Unlike ANC, captions describe the entire clip rather than localized events, so each test video contributes a single ``segment'' whose temporal extent coincides with the clip itself. We use MSR-VTT exclusively for evaluation, on its 2{,}990 test clips (${\sim}15$\,s each), to probe whether our model trained on ANC videos generalizes zero-shot to clips with a different caption distribution.

\subsection{Training and Evaluation}
We train PEEK on the ANC dataset \cite{krishnaDenseCaptioningEventsVideos2017a}, which provides videos with temporally grounded natural-language descriptions. Each annotated segment is treated as an independent training clip, decoded at a fixed frame rate of $2$~fps. During training, long sequences are capped and shorter sequences are zero-padded with an attention mask to allow batch training. No caption text, sentence boundaries, or external metadata are ever exposed to the Stage~2 model: the scorer sees only visual features and temporal positions.

We evaluate our trained selector on video captioning, on both ANC and MSR-VTT test sets, selecting $k \in \{1, 2, 4, 8\}$ frames that are fed to a downstream VLM. We compare PEEK against five training-free frame selection methods:
\begin{itemize}
    \item \textbf{Oracle} is the teacher, which sees the ground-truth captions; unusable at inference time, it estimates an approximate upper bound on the sampler's achievable performance.
    \item \textbf{Uniform} splits the (densely) sampled frames into $k$ equal temporal sub-segments and select the center frame of each.
    \item \textbf{Random} uses the same temporal sub-segments as Uniform and samples one frame at random from each sub-segment, using a fixed seed shared across all VLMs.
    \item \textbf{MaxInfo} \cite{liMaxInfoTrainingFreeKeyFrame2025} selects a diverse, high-information subset by applying a maximum-volume criterion to CLIP image embeddings; we use its fixed-cardinality mode with exactly $k$ selected frames.
    \item \textbf{CSTA} \cite{sonCSTACNNbasedSpatiotemporal2024} is originally a video summarization method predicting frame-importance scores; since our evaluation requires a fixed number of frames, we use only its scoring stage and select the highest-scoring frame in each of the $k$ temporal sub-segments.
    \item \textbf{PEEK} is our student model trained on ANC with stratified argmax selection.
\end{itemize}

We fix the parameters and seed for all evaluated downstream VLMs for fair comparison, and use the same candidate frames for all methods. We do not include PickNet~\cite{chenLessMorePicking2018} or LFS~\cite{chaoLFSLearnableFrame2026} in the quantitative comparison: neither has released code or checkpoints, PickNet leaves its reward-weighting coefficients and selection-length schedule unspecified, and LFS depends on captions re-distilled from a ${\sim}235$B-parameter teacher over an unreleased training set, so a faithful reimplementation is not possible. We instead add two reproducible content-aware, supervision-free selectors built on PEEK's own visual features (Section~\ref{sec:trainingfree}).

For each selected frame budget, we generate captions conditioned only on the $k$ chosen frames, in temporal order, and a short captioning prompt. Concretely, we pass the $k$ selected frames as a single multi-image input to the downstream VLM, followed by a one-sentence prompt; we do not use frame grids, timestamps, or explicit delimiters. We evaluate four downstream VLMs of various sizes: SmolVLM2-2.2B-Instruct~\cite{marafiotiSmolVLMRedefiningSmall2025}, Qwen2.5-VL-3B~\cite{baiQwen25VLTechnicalReport2025}, Qwen3.5-4B~\cite{qwen35blog} and Qwen2.5-VL-7B~\cite{baiQwen25VLTechnicalReport2025}. Prompt templates and model-specific generation settings are provided in the supplementary material. We report CIDEr~\cite{vedantamCIDErConsensusbasedImage2014}, BLEU-4~\cite{papineniBleuMethodAutomatic2002}, METEOR~\cite{banerjeeMETEORAutomaticMetric2005}, and ROUGE-L~\cite{linROUGEPackageAutomatic2004}, with all metrics shown on the same $\times 100$ scale. CIDEr remains the primary metric for discussion because it is the most commonly reported metric for video captioning.

\subsection{Results}
\label{sec:cap}

\subsubsection{ActivityNet Captions}

Table~\ref{tab:captioning-anc} reports the results on ActivityNet Captions. The results show that PEEK is the strongest of the evaluated query-free selectors on this benchmark, obtaining the best CIDEr in 14 out of 16 model/budget settings. The gains are most pronounced at \(k{=}1\), where PEEK improves over the strongest query-free baseline by \(+1.74\) CIDEr points for SmolVLM2-2.2B, \(+2.34\) for Qwen2.5-VL-3B, \(+2.18\) for Qwen3.5-4B, and \(+3.00\) for Qwen2.5-VL-7B. The same conclusion holds at \(k{=}2\), where PEEK is again best for all four VLMs, with gains ranging from \(+0.61\) to \(+1.75\) CIDEr points. Greedy decoding makes captions deterministic given the selected frames, so we test significance with paired per-video statistics (bootstrap with $10^4$ resamples and Wilcoxon signed-rank over per-video CIDEr; $n{=}17{,}504$ ANC segments, $2{,}990$ MSR-VTT videos). At \(k{=}1\), PEEK's improvement is significant against \emph{every} baseline for \emph{every} VLM on both datasets (32/32, all $p{<}0.01$). Against Uniform, the number of significant (dataset, VLM) configurations decays from 8/8 at \(k{=}1\) to 5/8, 3/8, and 1/8 at \(k{=}2,4,8\), quantifying how the gains concentrate in the low-budget regime.

Compared with the adaptive baselines, PEEK is consistently stronger in the low-budget regime. Random is close to Uniform but rarely improves substantially, so the gains are not explained by simply perturbing the center frame of each sub-segment; CSTA is generally below Uniform and PEEK, indicating that summarization-learned importance does not directly transfer to caption-oriented selection; and MaxInfo, the weakest method at \(k{=}1\) and inconsistent at larger budgets, shows that visual diversity alone is not equivalent to caption relevance.

At larger budgets, the advantage of PEEK becomes smaller but remains strong on ANC. PEEK is best in three out of four CIDEr settings at \(k{=}4\), losing only for Qwen2.5-VL-7B, where MaxInfo is higher. At \(k{=}8\), PEEK is also best in three out of four settings, losing only for Qwen3.5-4B, where Uniform is higher by \(0.11\) CIDEr points. These small reversals show that PEEK is not universally better than temporal coverage, but it is the most reliable query-free selector on ANC, especially when the frame budget is tight.
\begin{table*}[t]
\centering
\scriptsize
\setlength{\tabcolsep}{3pt}
\resizebox{\textwidth}{!}{%
\begin{tabular}{l l cccc cccc cccc cccc}
\toprule
VLM & Selector & \multicolumn{4}{c}{CIDEr} & \multicolumn{4}{c}{BLEU-4} & \multicolumn{4}{c}{METEOR} & \multicolumn{4}{c}{ROUGE-L} \\
\cmidrule(lr){3-6}\cmidrule(lr){7-10}\cmidrule(lr){11-14}\cmidrule(lr){15-18}
& & $k{=}1$ & $2$ & $4$ & $8$ & $k{=}1$ & $2$ & $4$ & $8$ & $k{=}1$ & $2$ & $4$ & $8$ & $k{=}1$ & $2$ & $4$ & $8$ \\
\midrule
\multirow{6}{*}{SmolVLM2-2.2B}
  & Oracle & \oracle{37.36} & \oracle{38.19} & \oracle{39.41} & \oracle{39.05} & \oracle{2.65} & \oracle{2.66} & \oracle{2.87} & \oracle{3.31} & \oracle{9.00} & \oracle{9.14} & \oracle{9.33} & \oracle{9.54} & \oracle{20.32} & \oracle{20.50} & \oracle{20.79} & \oracle{20.63} \\
  & Uniform & \underline{29.79} & \underline{31.23} & \underline{32.76} & \underline{33.85} & \underline{2.10} & \underline{2.19} & \underline{2.43} & \underline{2.94} & \underline{8.08} & \underline{8.31} & \underline{8.59} & \underline{8.95} & \underline{18.55} & \underline{19.00} & \underline{19.39} & \underline{19.52} \\
  & Random & 28.40 & 31.06 & 32.35 & 33.47 & 2.03 & 2.17 & 2.40 & 2.91 & 7.85 & 8.25 & 8.58 & 8.90 & 18.18 & 18.88 & 19.38 & 19.49 \\
  & CSTA~\cite{sonCSTACNNbasedSpatiotemporal2024} & 28.37 & 30.38 & 32.62 & 33.69 & 1.99 & 2.13 & 2.41 & 2.88 & 7.85 & 8.17 & 8.58 & 8.92 & 18.20 & 18.72 & 19.39 & 19.50 \\
  & MaxInfo~\cite{liMaxInfoTrainingFreeKeyFrame2025} & 27.07 & 29.49 & 31.91 & 33.06 & 1.92 & 2.07 & 2.32 & 2.79 & 7.47 & 8.05 & 8.47 & 8.81 & 17.55 & 18.60 & 19.23 & 19.31 \\
  & PEEK (ours) & \textbf{31.53}$^{***}$ & \textbf{32.98}$^{***}$ & \textbf{33.45}$^{***}$ & \textbf{34.33}$^{**}$ & \textbf{2.23} & \textbf{2.32} & \textbf{2.48} & \textbf{2.96} & \textbf{8.37} & \textbf{8.54} & \textbf{8.72} & \textbf{9.02} & \textbf{19.12} & \textbf{19.39} & \textbf{19.62} & \textbf{19.70} \\
\midrule
\multirow{6}{*}{Qwen2.5-VL-3B}
  & Oracle & \oracle{37.31} & \oracle{41.67} & \oracle{45.23} & \oracle{46.58} & \oracle{4.04} & \oracle{4.34} & \oracle{4.51} & \oracle{4.54} & \oracle{10.84} & \oracle{11.25} & \oracle{11.53} & \oracle{11.64} & \oracle{21.25} & \oracle{22.05} & \oracle{22.67} & \oracle{22.95} \\
  & Uniform & \underline{30.05} & \underline{35.36} & 39.51 & \underline{42.33} & \underline{3.19} & \underline{3.73} & \underline{4.00} & \underline{4.12} & \underline{9.72} & \underline{10.47} & 10.88 & \underline{11.12} & \underline{19.58} & \underline{20.81} & \underline{21.60} & \underline{22.11} \\
  & Random & 29.03 & 34.99 & \underline{39.54} & 41.91 & 3.17 & 3.67 & 3.96 & 4.11 & 9.54 & 10.39 & 10.84 & 11.07 & 19.35 & 20.71 & 21.59 & 22.05 \\
  & CSTA~\cite{sonCSTACNNbasedSpatiotemporal2024} & 28.56 & 34.53 & 39.16 & 42.06 & 3.09 & 3.60 & 3.88 & 4.09 & 9.48 & 10.29 & 10.79 & 11.07 & 19.21 & 20.60 & 21.46 & 22.02 \\
  & MaxInfo~\cite{liMaxInfoTrainingFreeKeyFrame2025} & 26.47 & 35.10 & 39.47 & 41.91 & 2.82 & 3.65 & 3.93 & 4.09 & 9.05 & 10.41 & \underline{10.89} & 11.08 & 18.46 & 20.71 & 21.59 & 22.08 \\
  & PEEK (ours) & \textbf{32.39}$^{***}$ & \textbf{37.02}$^{***}$ & \textbf{40.55}$^{**}$ & \textbf{42.42} & \textbf{3.45} & \textbf{3.80} & \textbf{4.06} & \textbf{4.18} & \textbf{10.13} & \textbf{10.61} & \textbf{10.93} & \textbf{11.13} & \textbf{20.21} & \textbf{21.06} & \textbf{21.77} & \textbf{22.12} \\
\midrule
\multirow{6}{*}{Qwen3.5-4B}
  & Oracle & \oracle{38.44} & \oracle{40.64} & \oracle{40.19} & \oracle{40.30} & \oracle{2.56} & \oracle{3.08} & \oracle{3.12} & \oracle{3.26} & \oracle{9.07} & \oracle{9.76} & \oracle{9.96} & \oracle{10.07} & \oracle{19.35} & \oracle{20.35} & \oracle{20.30} & \oracle{20.37} \\
  & Uniform & \underline{29.55} & \underline{33.35} & 34.66 & \textbf{35.42} & 1.93 & \underline{2.55} & 2.63 & \underline{2.76} & \underline{7.85} & \underline{8.72} & 9.12 & \underline{9.37} & \underline{17.28} & \underline{18.67} & 19.08 & \textbf{19.30} \\
  & Random & 28.94 & 32.93 & \underline{34.72} & 34.93 & \underline{1.94} & 2.50 & 2.64 & 2.69 & 7.69 & 8.70 & \underline{9.14} & 9.30 & 17.06 & 18.67 & \underline{19.09} & 19.25 \\
  & CSTA~\cite{sonCSTACNNbasedSpatiotemporal2024} & 28.51 & 33.47 & 34.54 & \underline{35.39} & 1.90 & 2.50 & \underline{2.66} & 2.74 & 7.63 & 8.69 & 9.10 & 9.33 & 16.95 & 18.59 & 19.02 & 19.26 \\
  & MaxInfo~\cite{liMaxInfoTrainingFreeKeyFrame2025} & 27.01 & 32.97 & 34.44 & 35.22 & 1.80 & 2.50 & 2.60 & 2.68 & 7.36 & 8.70 & 9.08 & 9.26 & 16.56 & 18.66 & 18.97 & 19.21 \\
  & PEEK (ours) & \textbf{31.73}$^{***}$ & \textbf{34.51}$^{***}$ & \textbf{34.93}$^{**}$ & 35.31 & \textbf{2.04} & \textbf{2.61} & \textbf{2.73} & \textbf{2.80} & \textbf{8.13} & \textbf{8.84} & \textbf{9.20} & \textbf{9.37} & \textbf{17.71} & \textbf{18.94} & \textbf{19.22} & \underline{19.29} \\
\midrule
\multirow{6}{*}{Qwen2.5-VL-7B}
  & Oracle & \oracle{36.60} & \oracle{40.10} & \oracle{38.05} & \oracle{33.88} & \oracle{2.60} & \oracle{2.64} & \oracle{2.27} & \oracle{1.92} & \oracle{9.24} & \oracle{9.64} & \oracle{9.18} & \oracle{8.58} & \oracle{19.59} & \oracle{21.12} & \oracle{20.84} & \oracle{19.69} \\
  & Uniform & \underline{28.54} & \underline{34.43} & 33.54 & 29.40 & 2.00 & \textbf{2.33} & 2.06 & \underline{1.71} & \underline{8.18} & \underline{8.91} & \underline{8.70} & \underline{8.01} & \underline{17.88} & \underline{19.84} & 19.96 & 18.70 \\
  & Random & 28.44 & 33.68 & \underline{33.55} & 29.58 & \underline{2.06} & 2.20 & \underline{2.07} & 1.68 & 8.08 & 8.80 & 8.69 & 8.01 & 17.79 & 19.67 & \underline{20.01} & 18.72 \\
  & CSTA~\cite{sonCSTACNNbasedSpatiotemporal2024} & 28.02 & 33.63 & 33.02 & \underline{29.77} & 2.02 & 2.25 & 2.03 & 1.69 & 8.03 & 8.78 & 8.62 & 7.99 & 17.71 & 19.63 & 19.89 & \underline{18.75} \\
  & MaxInfo~\cite{liMaxInfoTrainingFreeKeyFrame2025} & 26.44 & 33.56 & \textbf{33.92} & 29.43 & 1.99 & 2.23 & \textbf{2.12} & 1.67 & 7.76 & 8.88 & \textbf{8.73} & 7.97 & 17.20 & 19.73 & \textbf{20.11} & 18.72 \\
  & PEEK (ours) & \textbf{31.54}$^{***}$ & \textbf{35.04}$^{**}$ & 33.15 & \textbf{30.01} & \textbf{2.26} & \underline{2.31} & 2.03 & \textbf{1.71} & \textbf{8.58} & \textbf{9.05} & 8.67 & \textbf{8.06} & \textbf{18.54} & \textbf{20.06} & 19.92 & \textbf{18.75} \\
\bottomrule
\end{tabular}
}
\vspace{0.1pt}
\caption{ActivityNet Captions test captioning metrics with different downstream VLMs and frame budgets. PEEK uses the same ActivityNet-trained checkpoint for all downstream VLMs. Oracle scores frames against the ground-truth caption. \textbf{Bold} marks the best query-free method for each VLM, metric, and frame budget. \underline{Underline} is second-best. Stars on PEEK's CIDEr mark the significance of the paired per-video PEEK$-$Uniform difference (Wilcoxon signed-rank: $^{***}$\,$p{<}0.001$, $^{**}$\,$p{<}0.01$; see Section~\ref{sec:cap}).}
\label{tab:captioning-anc}
\end{table*}

\subsubsection{Zero-shot on MSR-VTT}

Table~\ref{tab:captioning-msrvtt} evaluates the same ActivityNet-trained selector on MSR-VTT, without retraining, testing whether PEEK learns a transferable visual relevance prior rather than an ANC-specific one. Transfer is strongest at \(k{=}1\): PEEK is the best query-free method for all four downstream VLMs on all reported metrics, improving CIDEr over the strongest query-free baseline by \(+2.68\) points for SmolVLM2-2.2B, \(+1.46\) for Qwen2.5-VL-3B, \(+2.26\) for Qwen3.5-4B, and \(+1.25\) for Qwen2.5-VL-7B. At \(k{=}2\), PEEK remains best in CIDEr for three of the four VLMs (Random is higher by \(0.19\) points for Qwen2.5-VL-3B) and is often best on the remaining metrics.

At \(k{=}4\) and \(k{=}8\) the comparison becomes mixed: PEEK stays close to the best query-free method throughout but wins CIDEr only for SmolVLM2-2.2B and Qwen2.5-VL-7B at \(k{=}8\), with Uniform, Random, or MaxInfo best elsewhere. The adaptive baselines are not uniformly stronger than Uniform either: MaxInfo is the weakest method at \(k{=}1\) despite explicitly optimizing diversity, and CSTA is often below Uniform. Overall, MSR-VTT sharpens the conclusion from ANC: PEEK transfers best in the low-budget regime, where semantic relevance matters most, while temporal coverage and diversity become increasingly competitive once several frames are available.

Several captioners are also non-monotonic in the number of frames: for Qwen2.5-VL-7B, CIDEr peaks at \(k{=}4\) and drops at \(k{=}8\) for every selector, including the Oracle (\(53.46 \rightarrow 47.08\)) despite its access to the reference caption, and SmolVLM2-2.2B degrades even more sharply at \(k{=}8\). These drops are not specific to PEEK, and caution against treating more frames as automatically better.

\begin{table*}[t]
\centering
\scriptsize
\setlength{\tabcolsep}{3pt}
\resizebox{\textwidth}{!}{%
\begin{tabular}{l l cccc cccc cccc cccc}
\toprule
VLM & Selector & \multicolumn{4}{c}{CIDEr} & \multicolumn{4}{c}{BLEU-4} & \multicolumn{4}{c}{METEOR} & \multicolumn{4}{c}{ROUGE-L} \\
\cmidrule(lr){3-6}\cmidrule(lr){7-10}\cmidrule(lr){11-14}\cmidrule(lr){15-18}
& & $k{=}1$ & $2$ & $4$ & $8$ & $k{=}1$ & $2$ & $4$ & $8$ & $k{=}1$ & $2$ & $4$ & $8$ & $k{=}1$ & $2$ & $4$ & $8$ \\
\midrule
\multirow{6}{*}{SmolVLM2-2.2B}
  & Oracle & \oracle{48.33} & \oracle{49.35} & \oracle{48.99} & \oracle{33.99} & \oracle{37.19} & \oracle{37.44} & \oracle{37.66} & \oracle{27.52} & \oracle{27.65} & \oracle{27.81} & \oracle{27.90} & \oracle{26.04} & \oracle{59.47} & \oracle{59.58} & \oracle{59.74} & \oracle{53.44} \\
  & Uniform & \underline{42.15} & 43.87 & \textbf{46.76} & 31.42 & \underline{32.26} & 34.06 & \underline{35.79} & \textbf{27.27} & \underline{25.53} & 26.02 & \underline{27.01} & 25.59 & \underline{56.01} & 57.11 & \underline{58.68} & \underline{53.12} \\
  & Random & 41.77 & \underline{44.53} & 46.18 & 32.52 & 31.93 & \underline{34.30} & 35.50 & 26.65 & 25.22 & \underline{26.11} & 26.92 & 25.51 & 55.85 & \underline{57.17} & 58.20 & 52.80 \\
  & CSTA~\cite{sonCSTACNNbasedSpatiotemporal2024} & 39.99 & 43.47 & 45.49 & \underline{32.99} & 31.24 & 33.91 & 35.43 & \underline{27.21} & 24.75 & 26.02 & 26.78 & \underline{25.72} & 54.99 & 57.06 & 58.29 & 52.90 \\
  & MaxInfo~\cite{liMaxInfoTrainingFreeKeyFrame2025} & 34.36 & 43.01 & 44.96 & 31.49 & 27.11 & 33.78 & 34.89 & 26.82 & 22.67 & 25.82 & 26.57 & 25.42 & 52.54 & 56.57 & 57.81 & 53.01 \\
  & PEEK (ours) & \textbf{44.83}$^{***}$ & \textbf{46.28}$^{***}$ & \underline{46.67} & \textbf{33.71} & \textbf{34.64} & \textbf{35.65} & \textbf{36.05} & 26.91 & \textbf{26.59} & \textbf{27.03} & \textbf{27.22} & \textbf{25.87} & \textbf{57.69} & \textbf{58.28} & \textbf{58.73} & \textbf{53.14} \\
\midrule
\multirow{6}{*}{Qwen2.5-VL-3B}
  & Oracle & \oracle{33.67} & \oracle{38.10} & \oracle{42.45} & \oracle{44.75} & \oracle{23.90} & \oracle{27.04} & \oracle{30.33} & \oracle{31.67} & \oracle{26.36} & \oracle{27.77} & \oracle{28.82} & \oracle{29.28} & \oracle{50.77} & \oracle{52.96} & \oracle{54.99} & \oracle{55.92} \\
  & Uniform & \underline{29.18} & 34.57 & \textbf{40.79} & 42.68 & \underline{21.74} & \underline{24.92} & \textbf{29.23} & 30.43 & \underline{24.50} & \underline{26.55} & \underline{28.12} & 28.79 & \underline{48.30} & \underline{51.17} & \textbf{54.16} & 55.29 \\
  & Random & 28.34 & \textbf{35.04} & 39.42 & \textbf{43.37} & 20.86 & 24.64 & 28.13 & \underline{31.07} & 24.13 & 26.32 & 27.91 & \underline{28.94} & 47.72 & 51.01 & 53.87 & \underline{55.51} \\
  & CSTA~\cite{sonCSTACNNbasedSpatiotemporal2024} & 28.20 & 34.19 & 40.34 & 43.09 & 20.73 & 24.71 & 28.60 & 30.62 & 23.97 & 26.41 & 27.90 & 28.90 & 47.51 & 50.80 & 53.89 & 55.24 \\
  & MaxInfo~\cite{liMaxInfoTrainingFreeKeyFrame2025} & 21.82 & 33.76 & 39.56 & \underline{43.18} & 16.88 & 24.18 & 28.37 & 30.79 & 21.85 & 26.28 & 27.98 & 28.93 & 44.45 & 50.91 & 53.83 & 55.43 \\
  & PEEK (ours) & \textbf{30.64}$^{**}$ & \underline{34.85} & \underline{40.36} & 42.94 & \textbf{22.60} & \textbf{25.28} & \underline{28.97} & \textbf{31.08} & \textbf{25.56} & \textbf{27.00} & \textbf{28.28} & \textbf{28.99} & \textbf{49.53} & \textbf{51.65} & \underline{54.08} & \textbf{55.51} \\
\midrule
\multirow{6}{*}{Qwen3.5-4B}
  & Oracle & \oracle{38.80} & \oracle{36.67} & \oracle{35.62} & \oracle{34.26} & \oracle{23.70} & \oracle{23.66} & \oracle{22.79} & \oracle{21.68} & \oracle{24.57} & \oracle{25.48} & \oracle{25.71} & \oracle{25.73} & \oracle{48.94} & \oracle{50.35} & \oracle{49.80} & \oracle{49.38} \\
  & Uniform & \underline{32.63} & 33.32 & 33.62 & 32.77 & \underline{20.74} & 21.58 & 21.32 & 20.88 & \underline{22.40} & \underline{24.29} & \textbf{25.27} & \underline{25.36} & \underline{45.96} & 48.40 & \textbf{49.00} & \underline{48.65} \\
  & Random & 32.21 & \underline{32.75} & 33.84 & 32.88 & 20.15 & 21.18 & 21.41 & \textbf{21.11} & 22.25 & 23.96 & 25.05 & 25.19 & 45.77 & 48.28 & 48.88 & 48.54 \\
  & CSTA~\cite{sonCSTACNNbasedSpatiotemporal2024} & 31.23 & 32.65 & 33.63 & 32.79 & 19.92 & 21.63 & \underline{21.62} & 20.64 & 21.97 & 24.23 & 24.92 & 25.20 & 45.26 & 48.41 & 48.66 & 48.39 \\
  & MaxInfo~\cite{liMaxInfoTrainingFreeKeyFrame2025} & 26.10 & 33.30 & \textbf{33.90} & \textbf{33.03} & 16.30 & \underline{21.73} & 21.56 & 20.82 & 19.83 & 24.27 & 24.94 & 25.26 & 42.78 & \underline{48.82} & \underline{48.89} & 48.49 \\
  & PEEK (ours) & \textbf{34.89}$^{***}$ & \textbf{33.92} & \underline{33.84} & \underline{32.89} & \textbf{21.62} & \textbf{22.37} & \textbf{21.62} & \underline{21.04} & \textbf{23.18} & \textbf{24.80} & \underline{25.09} & \textbf{25.37} & \textbf{46.89} & \textbf{49.24} & 48.82 & \textbf{48.65} \\
\midrule
\multirow{6}{*}{Qwen2.5-VL-7B}
  & Oracle & \oracle{36.63} & \oracle{52.42} & \oracle{53.46} & \oracle{47.08} & \oracle{25.30} & \oracle{37.98} & \oracle{39.63} & \oracle{36.17} & \oracle{25.88} & \oracle{28.55} & \oracle{28.53} & \oracle{26.81} & \oracle{51.54} & \oracle{59.43} & \oracle{59.84} & \oracle{56.90} \\
  & Uniform & 31.08 & \underline{48.57} & \textbf{51.42} & \underline{45.04} & 22.38 & \underline{35.48} & \textbf{39.10} & \underline{34.45} & \underline{23.84} & \underline{27.11} & \textbf{28.16} & \underline{26.13} & 48.55 & \underline{57.46} & \textbf{59.34} & \underline{56.00} \\
  & Random & \underline{31.91} & 47.13 & 48.95 & 42.10 & \underline{22.90} & 34.30 & 37.02 & 32.62 & 23.36 & 27.02 & 27.42 & 25.40 & \underline{49.01} & 57.00 & 58.47 & 54.82 \\
  & CSTA~\cite{sonCSTACNNbasedSpatiotemporal2024} & 31.54 & 46.85 & 49.12 & 41.71 & 22.38 & 34.61 & 37.31 & 32.53 & 23.18 & 26.79 & 27.41 & 25.23 & 48.61 & 56.96 & 58.43 & 54.48 \\
  & MaxInfo~\cite{liMaxInfoTrainingFreeKeyFrame2025} & 25.60 & 46.53 & 49.01 & 42.30 & 19.18 & 34.26 & 37.09 & 33.32 & 21.19 & 26.96 & 27.39 & 25.44 & 46.19 & 56.98 & 58.47 & 54.96 \\
  & PEEK (ours) & \textbf{33.16}$^{***}$ & \textbf{48.69} & \underline{50.99} & \textbf{45.29} & \textbf{23.55} & \textbf{36.00} & \underline{38.33} & \textbf{35.08} & \textbf{24.82} & \textbf{27.56} & \underline{27.94} & \textbf{26.34} & \textbf{50.01} & \textbf{58.02} & \underline{59.08} & \textbf{56.24} \\
\bottomrule
\end{tabular}
}
\vspace{0.1pt}
\caption{Zero-shot MSR-VTT test captioning metrics with different downstream VLMs and frame budgets. PEEK uses the same query-free ActivityNet-trained selector for all downstream VLMs. Oracle scores frames against the ground-truth caption. \textbf{Bold} marks the best query-free method for each VLM, metric, and frame budget. \underline{Underline} is second-best. Stars on PEEK's CIDEr mark the significance of the paired per-video PEEK$-$Uniform difference (Wilcoxon signed-rank: $^{***}$\,$p{<}0.001$, $^{**}$\,$p{<}0.01$; see Section~\ref{sec:cap}).}
\label{tab:captioning-msrvtt}
\end{table*}

\subsubsection{Content-aware training-free baselines}
\label{sec:trainingfree}

Diversity- or boundary-based heuristics are natural supervision-free competitors to a learned selector. We therefore evaluate two additional selectors that use the \emph{same} frozen MobileCLIP2-S0 features as PEEK's student, so all three methods see identical information: \emph{$k$-means diversity} (L2-normalized embeddings, $k$ clusters, medoid of each cluster, temporal order) and \emph{shot-boundary} (split at the $k{-}1$ largest consecutive-embedding changes, middle frame of each shot; at $k{=}1$ this reduces to the video midpoint, i.e.\ Uniform's frame). Table~\ref{tab:trainingfree} reports CIDEr at $k{=}1,2$.

$k$-means diversity is the strongest heuristic and beats Uniform, yet on ANC PEEK still obtains the best CIDEr in all 8 (VLM, $k$) settings against it (significant in 5/8 under the paired tests of Section~\ref{sec:cap}) and in all 8 settings against shot-boundary (all $p{<}10^{-4}$). Zero-shot on MSR-VTT, PEEK and $k$-means are statistically tied (PEEK significantly better in 1/6 comparisons, $k$-means in none), with both clearly above Uniform, while PEEK beats shot-boundary in all 6 comparisons ($p{\leq}0.003$). The learned selector's edge over the best diversity heuristic is therefore in-domain, and unlike the heuristics PEEK also outputs per-frame relevance scores that downstream systems can reuse.

\begin{table}[t]
  \centering
  \small
  \setlength{\tabcolsep}{3pt}
  \adjustbox{max width=\linewidth}{%
  \begin{tabular}{l l cccc cccc}
    \toprule
    Dataset & VLM & \multicolumn{4}{c}{$k{=}1$} & \multicolumn{4}{c}{$k{=}2$} \\
    \cmidrule(lr){3-6}\cmidrule(lr){7-10}
    & & PEEK & $k$-means & Shot-bnd. & Uniform & PEEK & $k$-means & Shot-bnd. & Uniform \\
    \midrule
    \multirow{4}{*}{ANC}
      & SmolVLM2-2.2B & \textbf{31.53} & 31.09 & 29.79 & 29.79 & \textbf{32.98} & 30.94 & 30.06 & 31.23 \\
      & Qwen2.5-VL-3B & \textbf{32.39} & 31.55 & 30.05 & 30.05 & \textbf{37.02} & 35.50 & 34.62 & 35.36 \\
      & Qwen3.5-4B    & \textbf{31.73} & 30.93 & 29.55 & 29.55 & \textbf{34.51} & 34.07 & 32.81 & 33.35 \\
      & Qwen2.5-VL-7B & \textbf{31.54} & 30.44 & 28.54 & 28.54 & \textbf{35.04} & 34.52 & 33.41 & 34.43 \\
    \midrule
    \multirow{3}{*}{MSR-VTT}
      & SmolVLM2-2.2B & 44.83 & \textbf{45.06} & 42.15 & 42.15 & \textbf{46.28} & 43.74 & 43.34 & 43.87 \\
      & Qwen2.5-VL-3B & 30.64 & \textbf{30.94} & 29.18 & 29.18 & 34.85 & \textbf{35.48} & 32.00 & 34.57 \\
      & Qwen3.5-4B    & \textbf{34.89} & 34.88 & 32.63 & 32.63 & 33.92 & \textbf{34.32} & 31.83 & 33.32 \\
    \bottomrule
  \end{tabular}%
  }
  \vspace{4pt}
  \caption{CIDEr for PEEK versus two content-aware, supervision-free selectors computed on the same MobileCLIP2-S0 features. Shot-boundary at $k{=}1$ coincides with Uniform's midpoint frame. Qwen2.5-VL-7B is omitted on MSR-VTT because its baseline captions were generated in an earlier evaluation wave whose generation drift ($\approx{+}1.6$ CIDEr offset on identical frames) invalidates cross-run pairing; its within-run comparison is consistent with the reported tie.}
  \label{tab:trainingfree}
\end{table}

\subsubsection{Reference-free evaluation}
\label{sec:llmjudge}

Reference-based metrics can penalize correct captions that differ from the reference, so we complement them with a reference-free study at $k{=}1$ (SmolVLM2-2.2B captioner, $n{=}500$ videos per dataset). An LLM judge (Claude Sonnet~5) receives up to 20 frames spanning the clip --- including both selectors' picked frames, unlabeled --- and never sees the ground-truth references; it either compares the two captions in randomized order or blindly scores each caption's factual consistency out of 10. On ANC the judge prefers PEEK's caption over Uniform's in $61.9\%$ of decided cases (185 vs.\ 114, sign test $p{=}4.8\times10^{-5}$) and scores PEEK higher on factual consistency ($6.01$ vs.\ $5.58$, Wilcoxon $p{=}0.019$). Zero-shot on MSR-VTT, the two selectors are statistically indistinguishable on both instruments ($p{=}0.85$ and $p{=}0.36$), consistent with the smaller CIDEr margins there. The ANC preference is stable under a smaller 8-frame judge context ($59.3\%$, $p{=}0.002$), so it is not a consequence of context size. The reference-based gains at $k{=}1$ therefore correspond to captions that are genuinely preferred and more factually consistent when judged against the video itself.

\subsection{Efficiency}
\label{sec:efficiency}

Table~\ref{tab:efficiency} reports selection and end-to-end captioning time on the full ANC evaluation split. Uniform and Random have negligible selection cost, while all content-aware methods add a scoring pass over the candidate frames: PEEK scores all 17,505 segments in 1h44m of GPU time (\(0.36\)s per segment), against 21h58m (\(4.52\)s) for CSTA, 71h04m (\(14.62\)s) for MaxInfo, and 9h52m (\(2.03\)s) for the Oracle, which is in any case not deployable because it uses the ground-truth caption. Reused over the full \(k\in\{1,2,4,8\}\) captioning pipeline, PEEK increases total GPU time by only \(5.2\%\) over Uniform, against \(65.4\%\) for CSTA, \(211.9\%\) for MaxInfo, and \(29.4\%\) for the Oracle. PEEK is not free, but it is far cheaper than the other content-aware selectors --- lightweight enough to serve as a practical preprocessing stage while recovering part of the Oracle's caption-relevance signal.

\begin{table}[t]
\centering
\small
\setlength{\tabcolsep}{5pt}
\begin{tabular}{l c c c c}
\toprule
Selector & Text at inference & Selector time & Per segment & Full pipeline \\
\midrule
Uniform/Random & No & negligible & -- & 33h35m \\
PEEK (ours) & No & 1h44m {\scriptsize(26m)} & \textbf{0.36s} & 35h20m {\scriptsize(+5.2\%)} \\
CSTA~\cite{sonCSTACNNbasedSpatiotemporal2024} & No & 21h58m {\scriptsize(5h31m)} & 4.52s & 55h33m {\scriptsize(+65.4\%)} \\
MaxInfo~\cite{liMaxInfoTrainingFreeKeyFrame2025} & No & 71h04m {\scriptsize(17h49m)} & 14.62s & 104h44m {\scriptsize(+211.9\%)} \\
Oracle & Yes & 9h52m {\scriptsize(2h28m)} & 2.03s & 43h27m {\scriptsize(+29.4\%)} \\
\bottomrule
\end{tabular}
\vspace{6pt}
\caption{
Selection and end-to-end captioning time on the full ActivityNet Captions evaluation split with 17,505 segments, with SmolVLM2-2.2B-Instruct. Timings are measured on $4\times$NVIDIA A10G GPUs. We report total GPU time, with 4-GPU wall-clock estimates in parentheses. The full pipeline evaluates $k\in\{1,2,4,8\}$.
}
\label{tab:efficiency}
\end{table}

\subsection{Qualitative analysis}
\label{sec:qualitative}

\begin{figure*}[t]
  \centering
  \includegraphics[width=\linewidth]{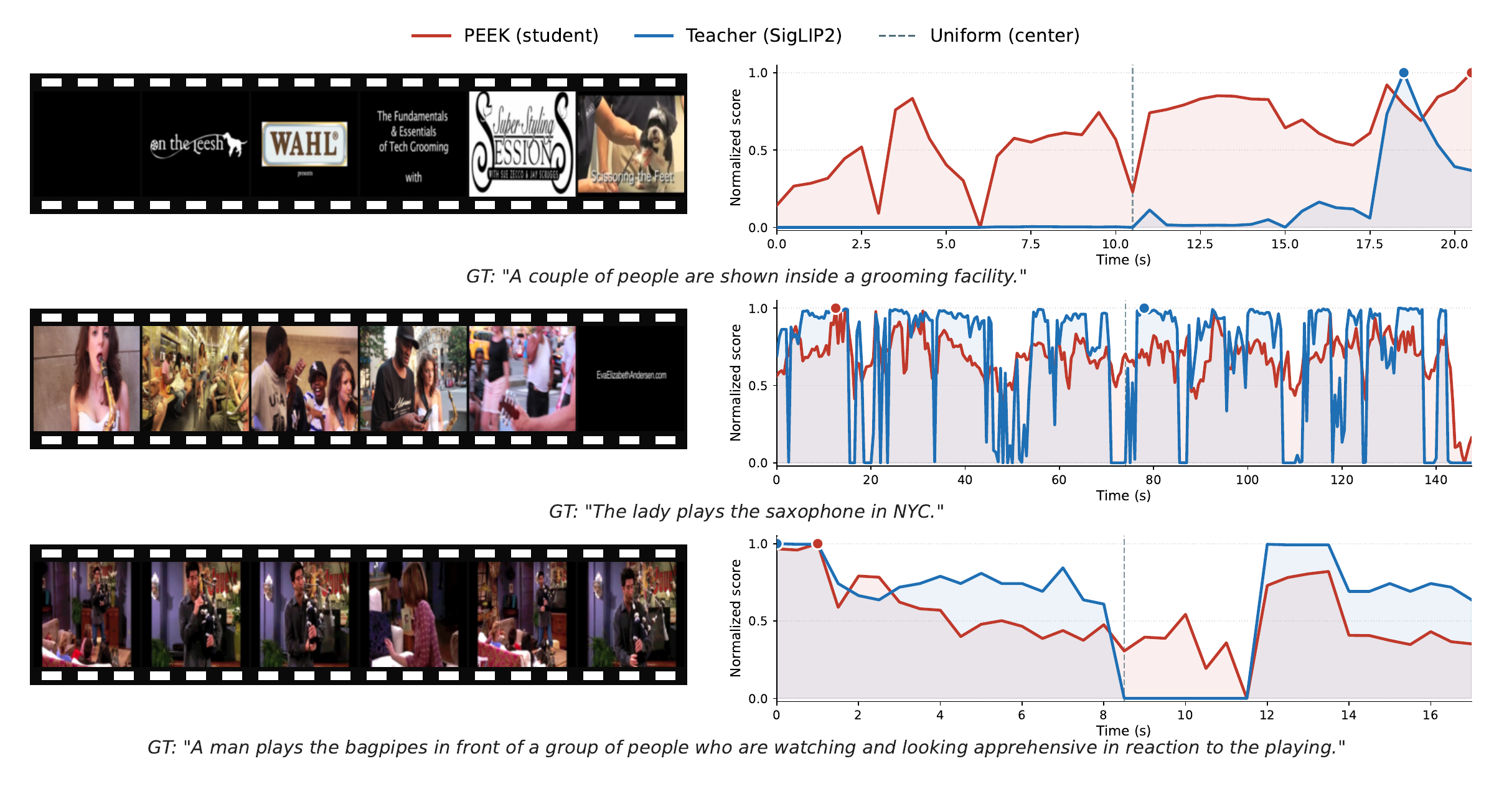}
  \caption{%
    Per-frame relevance scores on three ActivityNet Captions test segments. Curves are min--max normalized per video and markers indicate the argmax frame for each method. PEEK (red) and SigLIP2 (blue) agree on the global temporal structure but disagree locally, and their top-frame choices differ.%
  }
  \label{fig:qual_score_curves}
\end{figure*}

Figure~\ref{fig:qual_score_curves} compares PEEK and SigLIP2 scores on ANC test segments. The two methods agree on global salient regions but often differ locally, with PEEK producing smoother temporal profiles than the frame-wise Oracle. Figure~\ref{fig:qual_selection} shows one such case: both PEEK and the Oracle identify the bagpipes, while the uniform center frame misses the instrument. Additional examples are provided in the supplementary material.

We also verify quantitatively that the student learns caption-conditioned relevance rather than a generic photographic-saliency proxy. Over 300 random ANC test segments, the per-segment Spearman correlation between the student's scores and the teacher's caption-conditioned relevance is $\rho{=}0.25$ (Wilcoxon vs.\ zero across segments, $p{=}2.5\times10^{-21}$), whereas its correlation with pixel-only proxies is indistinguishable from zero ($\rho{=}0.01$ with Laplacian sharpness, $\rho{=}0.00$ with grayscale contrast); paired per segment, the teacher correlation exceeds both proxies by $\Delta\rho{\approx}0.24$ ($p{\approx}10^{-12}$), and the teacher's own scores are likewise uncorrelated with these proxies. The distilled signal is therefore semantic rather than photometric.

\subsection{Ablations}
\label{sec:ablations}

We ablate two design choices of PEEK on MSR-VTT with Qwen2.5-VL-3B: the inference-time conversion of frame scores into a fixed-size frame set, and the training loss used to distill the teacher ranking.

First, we compare raw top-$k$ selection, which takes the $k$ highest-scoring frames globally, with stratified argmax, which selects the highest-scoring frame within each of $k$ equal temporal bins. Table~\ref{tab:msrvtt-topk-vs-stratified} shows that stratified argmax consistently improves over raw top-$k$ across all metrics and budgets. This confirms that learned relevance scores alone are not sufficient: temporal coverage remains important to avoid selecting near-duplicate frames around the same high-score region.

\begin{table}[t]
  \centering
  \small
  \setlength{\tabcolsep}{3pt}
  \adjustbox{max width=\linewidth}{%
  \begin{tabular}{l ccc ccc ccc ccc}
    \toprule
    Selection strategy & \multicolumn{3}{c}{CIDEr} & \multicolumn{3}{c}{BLEU-4} & \multicolumn{3}{c}{METEOR} & \multicolumn{3}{c}{ROUGE-L} \\
    \cmidrule(lr){2-4}\cmidrule(lr){5-7}\cmidrule(lr){8-10}\cmidrule(lr){11-13}
    & $k{=}2$ & $4$ & $8$ & $k{=}2$ & $4$ & $8$ & $k{=}2$ & $4$ & $8$ & $k{=}2$ & $4$ & $8$ \\
    \midrule
    Raw top-$k$ & 32.79 & 36.52 & 39.47 & 24.09 & 26.88 & 28.96 & 26.44 & 27.38 & 28.26 & 50.51 & 52.51 & 54.02 \\
    Stratified argmax & \textbf{34.85} & \textbf{40.36} & \textbf{42.94} & \textbf{25.28} & \textbf{28.97} & \textbf{31.08} & \textbf{27.00} & \textbf{28.28} & \textbf{28.99} & \textbf{51.65} & \textbf{54.08} & \textbf{55.51} \\
    \bottomrule
  \end{tabular}%
  }
  \vspace{4pt}
  \caption{MSR-VTT metrics with Qwen2.5-VL-3B when converting PEEK scores into selected frames using raw top-$k$ or stratified argmax.}
  \label{tab:msrvtt-topk-vs-stratified}
\end{table}

Second, we compare the ListMLE loss with a pointwise MSE loss combined with a pairwise ranking loss. As shown in Table~\ref{tab:msrvtt-loss-ablation}, ListMLE improves all metrics at both $k{=}1$ and $k{=}2$, with the largest CIDEr gain at $k{=}1$. This motivates our use of a listwise objective. Implementation details can be found in the supplementary material.

\begin{table}[t]
  \centering
  \small
  \setlength{\tabcolsep}{3pt}
  \adjustbox{max width=\linewidth}{%
  \begin{tabular}{l cc cc cc cc}
    \toprule
    Loss & \multicolumn{2}{c}{CIDEr} & \multicolumn{2}{c}{BLEU-4} & \multicolumn{2}{c}{METEOR} & \multicolumn{2}{c}{ROUGE-L} \\
    \cmidrule(lr){2-3}\cmidrule(lr){4-5}\cmidrule(lr){6-7}\cmidrule(lr){8-9}
    & $k{=}1$ & $2$ & $k{=}1$ & $2$ & $k{=}1$ & $2$ & $k{=}1$ & $2$ \\
    \midrule
    MSE + pairwise & 29.46 & 34.49 & 21.79 & 25.06 & 25.14 & 26.73 & 48.93 & 51.54 \\
    ListMLE & \textbf{30.64} & \textbf{34.85} & \textbf{22.60} & \textbf{25.28} & \textbf{25.56} & \textbf{27.00} & \textbf{49.53} & \textbf{51.65} \\
    \bottomrule
  \end{tabular}%
  }
  \vspace{4pt}
  \caption{MSR-VTT metrics with Qwen2.5-VL-3B for PEEK trained with ListMLE or with an MSE + pairwise loss.}
  \label{tab:msrvtt-loss-ablation}
\end{table}

\section{Discussion and limitations}

Learned frame selection is most useful when the visual budget is tight: across both benchmarks, PEEK is the best query-free selector in all one-frame CIDEr settings and in most two-frame settings. This supports that part of the caption-conditioned relevance signal produced by an Oracle teacher can be recovered from visual evidence alone. The comparisons with CSTA, MaxInfo, and the diversity heuristics of Section~\ref{sec:trainingfree} show that PEEK is not generic summarization or visual diversity, but a caption-oriented notion of relevance that pays off when only one or two frames reach the captioner.

Learned selection is not universally preferable to uniform sampling, however. Uniform remains a strong baseline when several frames can be forwarded, most visibly on MSR-VTT at \(k{=}4\), where it often obtains the best CIDEr. The likely reason is the evaluation setting: ANC segments and MSR-VTT clips are short, so a few uniformly spaced frames often cover the main event, and as the budget grows, temporal coverage and diversity matter more than picking the single most relevant frame.

Another limitation is that the teacher signal is derived from ground-truth captions. This ties the learned notion of relevance to reference-caption alignment rather than to all visually meaningful events: a frame that supports a correct but non-reference caption may receive a weak teacher score. The same dependence affects reference-based captioning metrics, which can penalize correct captions that differ from the reference and can behave non-monotonically as more visual context is added. The reference-free evaluation of Section~\ref{sec:llmjudge} mitigates this dependence, and extending the analysis to adaptive frame budgets and human judgments remains future work. 

A final limitation is that our evaluation is restricted to short-caption generation. Long-form video captioning may require preserving multiple events, fine-grained temporal order, and details that a single caption-conditioned relevance ranking does not capture, and selecting only the most caption-aligned frames could overemphasize the dominant event at the expense of secondary cues. As a preliminary probe, we evaluated PEEK against Uniform on 100 videos of 8--10 minutes from CapRiCorn-1K~\cite{chenCapRiCorn1K2026} with Qwen2.5-VL-3B: PEEK improves annotated-event coverage at every tested budget ($+2.9$ points at $k{=}8$, $p{<}10^{-4}$; $+2.4$ at $k{=}32$, $p{=}0.002$; paired Wilcoxon), despite being trained on short ANC segments only, while caption accuracy is unchanged; a full long-form study remains future work. Finally, although PEEK is query-free by design, tasks such as video question answering or retrieval may benefit from task- or query-specific selection; PEEK could serve there as a lightweight first-stage selector or a transferable initialization, but evaluating this requires dedicated experiments. Extending the distillation framework to longer descriptions, adaptive frame budgets, and query-conditioned supervision is therefore an important direction for future work.

\section{Conclusion}

We introduced PEEK, a query-free frame selector for video captioning trained by distilling caption-conditioned relevance rankings from an Oracle teacher into a lightweight temporal model, making part of the caption-aware selection signal usable at inference time, when no caption is available. Across ActivityNet Captions and MSR-VTT, PEEK is the strongest of the evaluated query-free selectors, obtaining the best CIDEr in all one-frame settings, in seven out of eight two-frame settings, and in 23 out of 32 CIDEr comparisons overall. The gains are clearest when the frame budget is tight: zero-shot transfer is strongest at \(k{=}1\) and \(k{=}2\), while at \(k{=}4\) and \(k{=}8\) Uniform, Random, or MaxInfo occasionally perform slightly better. Caption-relevance distillation therefore pays off most when only a few frames can be used, while temporal coverage and diversity remain preferable at larger budgets.

PEEK also provides a favorable efficiency trade-off: much faster than CSTA~\cite{sonCSTACNNbasedSpatiotemporal2024} and MaxInfo~\cite{liMaxInfoTrainingFreeKeyFrame2025} while consistently outperforming them in the low-frame regime, it is a practical selector for efficient video captioning, thumbnail choice, and preview-frame selection.

\section{Acknowledgments}

This work was granted access to the HPC resources of IDRIS under the allocation 20XX-[AD011017404] made by GENCI.
\FloatBarrier

\bibliography{egbib}
\end{document}